\documentclass[runningheads]{llncs}
\usepackage[T1]{fontenc}
\usepackage{graphicx}
\usepackage{amsmath,amssymb}
\usepackage{booktabs,multirow,tabularx,array}
\usepackage{enumitem}
\usepackage{adjustbox}
\usepackage{microtype}
\usepackage{url}
\graphicspath{{figures/}}

\begin{document}
\title{Uncertainty Gated Quantum Kernel Residuals\\for Face Verification}
\titlerunning{Uncertainty-Gated Quantum Kernel Residuals}
\author{Krishna Bhatia\inst{1}, Kavita Choudhary\inst{2}, Ram Prakash Sharma\inst{3}, Pruthwik Mishra\inst{2} \and Praveen Kumar Chandaliya\inst{2}}
\authorrunning{Bhatia et. al}
\institute{QuantumAI Lab, Fractal Analytics, Mumbai, India\\\email{krishna.bhatia@fractal.ai}
\and  Sardar Vallabhbhai National Institute of Technology, Surat, India\\\email{ds25ai002@aid.svnit.ac.in,pruthwikmishra@aid.svnit.ac.in,pkc@aid.svnit.ac.in} \and National Institute of Technology, Hamirpur, India\\\email{ram.sharma@nith.ac.in}}
\maketitle

\begin{abstract}
Quantum-kernel SVMs are attractive near-term classifiers, but modern face embeddings already exhibit a strong cosine geometry and must be aggressively compressed to fit a small qubit budget. We study a more conservative role for a quantum kernel: an \emph{uncertainty-gated residual specialist} that is invoked only for face pairs for which the strong cosine verifier is ambiguous. A fixed ArcFace-style backbone supplies 512-D normalized embeddings; pair features are compressed to 8 or 12 dimensions and evaluated by an analytic-statevector quantum fidelity kernel. We compare cosine similarity, full and qubit-budget-matched RBF SVMs, standalone QkSVM, a random residual, a classically gated residual, and the quantum-gated residual across LFW, CASIA-WebFace, CFP-FP, and VGGFace2 protocols. Standalone budgeted kernels remain weaker than the full embedding geometry, but selective residual routing substantially reduces this failure. On the hard CFP-FP protocol, cosine obtains 0.816 AUC and 0.206 TAR at FAR=$10^{-2}$, standalone QkSVM 0.677/0.054, and the gated quantum residual 0.824/0.226. The identically routed classical residual achieves the same AUC of 0.824 and TAR of 0.223, demonstrating that the observed contribution is attributable to the hybrid uncertainty-routing strategy rather than to quantum advantage. On VGGFace2, where cosine is near ceiling (0.992 AUC), gating preserves 0.990 AUC while standalone QkSVM falls to 0.961. Across the four datasets, the gated quantum residual improves AUC over the standalone QkSVM by approximately 0.003, 0.007, 0.147, and 0.029, respectively. These results establish a practical design principle for quantum-assisted biometrics: when a compact quantum verifier is globally weaker, it should be treated as a selectively routed specialist and evaluated against an identically gated classical control.
\end{abstract}

\section{Introduction}
Face verification is dominated by highly discriminative deep embeddings. FaceNet~\cite{schroff2015facenet}, CosFace~\cite{wang2018cosface}, and ArcFace~\cite{deng2019arcface} make normalized cosine similarity an unusually strong decision rule. This creates a challenge for quantum machine learning (QML): a quantum verifier does not start from raw pixels, but from a representation whose geometry has already been optimized for a simple classical metric. Quantum kernels are nevertheless appealing for near-term learning because the quantum processor is used to estimate similarities while the support-vector optimization remains classical~\cite{schuld2019featurehilbert,havlicek2019supervised}. The difficulty is that a 512-D face representation or pair feature must be compressed to a small qubit count, and the kernel matrix requires quadratic numbers of pairwise evaluations. Under such constraints, a standalone QkSVM may be a poor replacement for cosine even if its feature map is nonlinear.

This paper investigates a systems question rather than a replacement question: \emph{can a budget-constrained quantum kernel serve as a residual specialist for uncertain face pairs?} The proposed architecture retains cosine similarity as the default score and routes only ambiguous pairs to a compact residual verifier. We evaluate this design against three critical controls: a full RBF classifier, a qubit-matched RBF classifier, and a gated classical residual with identical routing. A random residual further tests whether any gain arises merely from perturbing difficult pairs. The experiments span near-saturated benchmarks and a deliberately challenging cross-pose setting. LFW and CASIA-WebFace measure the limited headroom of a strong embedding, VGGFace2 provides an additional near-ceiling reference, and CFP-FP tests severe frontal-to-profile variation. We report AUC, EER, and TAR at fixed FAR to assess both overall discrimination and operational verification performance. Our contributions are:
\begin{itemize}[leftmargin=*]
\item We formulate \textbf{uncertainty-gated residual verification} as a conservative integration strategy for a compact QkSVM in a modern face-verification pipeline.
\item We provide \textbf{matched causal controls}: full RBF, qubit-budgeted RBF, random residual, and identically gated classical residual, preventing routing gains from being misattributed to the quantum kernel.
\item We show that gating is most useful on CFP-FP, where the gated quantum residual improves from 0.816 to 0.824 AUC over cosine and from 0.677 to 0.824 over standalone QkSVM, while preserving near-ceiling performance on VGGFace2.
\item We show that the gated classical residual reaches the same CFP-FP AUC, so the defensible conclusion is \emph{hybrid routing effectiveness}, not quantum advantage.
\item We quantify the qubit/kernel-computation bottleneck and derive deployment-oriented guidance for using weak quantum specialists inside strong biometric systems.
\end{itemize}

\section{Related Work}
\subsection{Deep Face Verification}
Modern face recognition represents each aligned face by an embedding in which same-identity samples are close and different identities are separated. FaceNet introduced triplet-based deep metric learning~\cite{schroff2015facenet}; SphereFace, AM-Softmax, CosFace, and ArcFace progressively sharpened angular decision margins~\cite{liu2017sphereface,wang2018amsoftmax,wang2018cosface,deng2019arcface}. MagFace additionally links representation magnitude to recognizability~\cite{meng2021magface}. These methods motivate our use of cosine similarity as the primary system rather than a weak baseline.

\subsection{Quantum Kernels and Near-Term Constraints}
Quantum feature maps encode classical vectors into quantum states so that fidelity or projected similarities define kernels~\cite{schuld2019featurehilbert,havlicek2019supervised}. Kernel methods avoid training a deep variational circuit but require many quantum evaluations. For $n$ training points and $m$ test points, a precomputed kernel requires $O(n^2 + mn)$ evaluations. High-dimensional image representations also have to be compressed to the available qubits. These effects are particularly relevant in biometrics because operational protocols can include many pairs and low-FAR evaluation is sensitive to score perturbations. Trainable quantum embeddings and data re-uploading can improve expressivity~\cite{hubregtsen2022training,perezsalinas2020datareuploading}, but near-term models remain vulnerable to trainability and concentration phenomena~\cite{mcclean2018barren}. We therefore intentionally use a shallow analytic kernel and focus on how it is integrated into the face system.

\subsection{Selective and Hybrid Decision Systems}
Selective prediction and mixture-of-experts systems exploit the fact that different models are useful on different samples. Our use case is asymmetric: cosine is cheap and strong globally; the quantum kernel is expensive and weak globally but may still encode complementary nonlinear structure on ambiguous pairs. The gate therefore acts as a risk-control mechanism. The matched classical residual is essential because any benefit from selective routing must be separated from benefit specific to the quantum kernel.

\section{Methodology}
\subsection{Fixed Face Representation and Pair Features}
A fixed ArcFace-style InsightFace backbone produces 512-D $\ell_2$-normalized embeddings $e_1,e_2$. Cosine similarity is $
s_c=e_1^\top e_2$. Learning-based residual verifiers use the order-invariant pair feature $x=|e_1-e_2|$.
PCA and standardization are fitted only on the training partition of each fold. The compact verifiers receive $q\in\{8,12\}$ dimensions; the full RBF control receives the available full pair representation.
\subsection{Quantum-Kernel Specialist}
The compact quantum verifier maps the standardized $q$-dimensional feature vector to angles in $[-\pi,\pi]$ and encodes them using $R_Y$ rotations. A shallow ring feature map employs nearest-neighbour entanglement with wrap-around coupling. The resulting fidelity kernel is
\begin{equation}
k(\mathbf{x},\mathbf{y})
=
\left|
\langle 0|U(\mathbf{x})^\dagger U(\mathbf{y})|0\rangle
\right|^2.
\end{equation}
Kernels are evaluated analytically with PennyLane~\cite{bergholm2018pennylane} (shots$=0$), eliminating sampling noise and isolating the effects of representation and uncertainty-based routing. The resulting kernel matrices are then provided to a classical SVM using a precomputed-kernel formulation.

\subsection{Uncertainty-Gated Residual Integration}
The primary verifier remains cosine. The gate identifies a validation-calibrated ambiguous region around the cosine decision boundary. Pairs outside this region keep their cosine score. Pairs inside the region receive a residual correction from a specialist score. In abstract form,
\begin{equation}
s_{\mathrm{gate}}=s_c+g(s_c)\,r(x), \qquad g(s_c)\in\{0,1\},
\end{equation}
where $g=1$ only for ambiguous pairs and $r(x)$ is the validation-calibrated residual produced by the specialist. The same routing mask and calibration procedure are applied to the quantum and classical residual controls; therefore differences between them cannot be explained by a more favorable gate.

We evaluate four residual variants: (i) random residual, which tests whether arbitrary perturbation of ambiguous scores helps; (ii) matched classical residual using the qubit-dimensional RBF model; (iii) quantum residual using QkSVM; and (iv) the original cosine score with no residual. The design intentionally treats the specialist as complementary rather than as a universal replacement.

\begin{figure}[t]
\centering
\includegraphics[width=\textwidth]{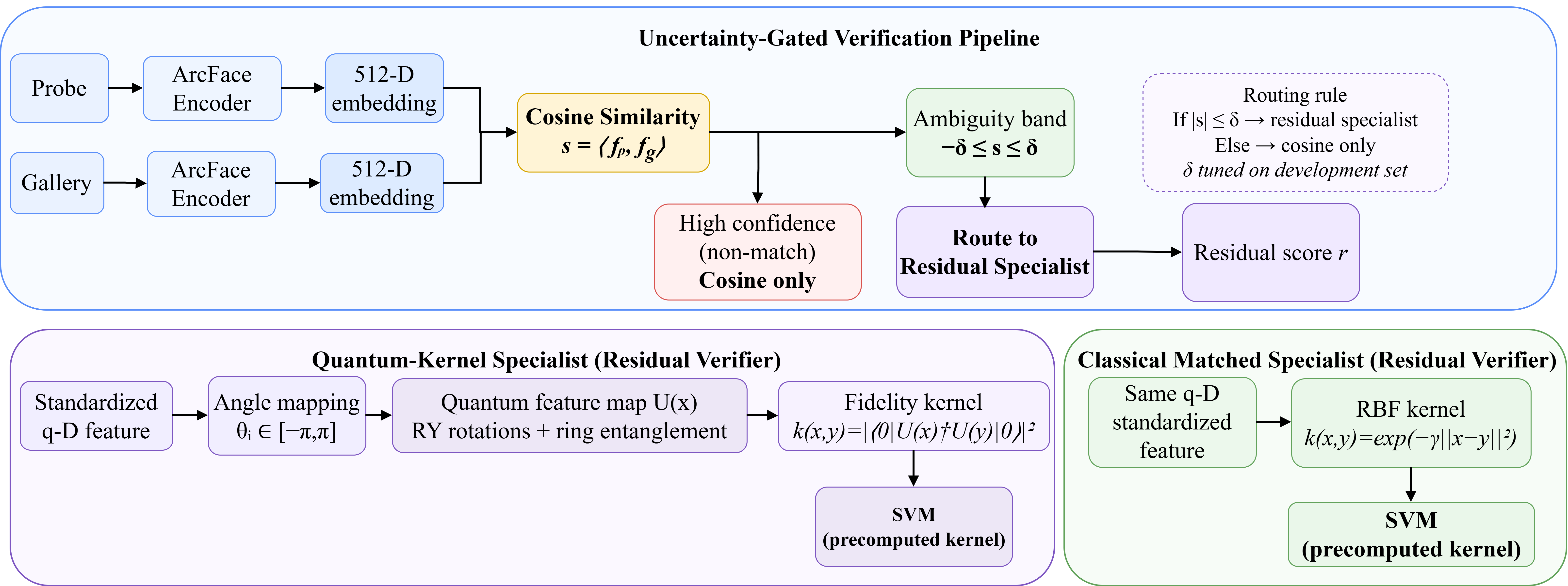}
\caption{Uncertainty-gated face verification. Cosine similarity handles high-confidence pairs, while ambiguous pairs are routed to either a quantum-kernel or matched classical residual specialist under the same gating rule.}
\label{fig:pipeline}
\end{figure}


\subsection{Kernel Cost}
For each fold the QkSVM requires a symmetric training kernel and a test-to-training kernel $O(n_{\mathrm{train}}^2+n_{\mathrm{test}}n_{\mathrm{train}})$. The two completed near-term configurations use $(q,n_{\mathrm{train}})=(8,300)$ and $(12,600)$ balanced training pairs, corresponding to approximately $2.25\times10^5$ and $5.40\times10^5$ circuit evaluations per fold under the reported setup. This cost motivates the selective use of the quantum specialist.

\section{Experimental Protocol}

\subsection{Datasets}

We evaluate on four face-verification benchmarks spanning standard unconstrained matching and challenging pose variation. LFW follows the standard 6,000-pair, 10-fold unconstrained verification protocol~\cite{huang2007lfw}; CFP-FP evaluates frontal-to-profile verification and provides the strongest pose shift in the study~\cite{sengupta2016cfp}; CASIA-WebFace provides a larger identity pool for the multi-seed development and evaluation study~\cite{yi2014casiawebface}; and VGGFace2 provides substantial pose and age variation as an external verification setting~\cite{cao2018vggface2}. The experiments were conducted in two stages: an initial 8/12-qubit study on LFW/CASIA-WebFace and a subsequent controlled CFP-FP/VGGFace2 study incorporating kernel-target alignment and uncertainty-gated residual verification. Accordingly, comparisons are made only within the same dataset and experimental protocol, and raw scores across stages are not interpreted as direct measures of dataset difficulty.


\subsection{Metrics and Controls}

We report ROC AUC, EER, and TAR at $\mathrm{FAR}=10^{-2}$ for the controlled cross-dataset analysis, together with the completed low-FAR LFW evaluation. All thresholds and calibration parameters are estimated exclusively from the training/development folds and then applied unchanged to the corresponding held-out folds. The evaluation includes cosine similarity, full-dimensional RBF, qubit-dimensional matched RBF, standalone QkSVM, gated random residual, gated classical residual, and gated quantum residual as the principal controls and comparison methods.


\begin{table}[t]
\centering
\caption{Approximate QkSVM kernel cost per fold.}
\label{tab:cost}
\begin{tabular}{ccccc}
\toprule
\textbf{$q$} & \textbf{$n_{\rm train}$} & \textbf{$K_{\rm train}$} & \textbf{$K_{\rm test}$} & \textbf{evals}\\
\midrule
8 & 300 & $300\times300$ & $600\times300$ & $2.25\times10^5$\\
12 & 600 & $600\times600$ & $600\times600$ & $5.40\times10^5$\\
\bottomrule
\end{tabular}
\end{table}

\section{Results}

\subsection{Qubit Budget: Capacity Versus Kernel Cost}
The completed LFW study provides a direct view of the compactness bottleneck before residual routing is introduced. Increasing the encoding from $q=8$ to $q=12$ and doubling the balanced training subset from 300 to 600 pairs improves the stringent LFW TAR at FAR=$10^{-3}$ from approximately 0.14 to 0.57, while the classical face representation remains near the accuracy ceiling. The result shows that the quantum kernel is not entirely insensitive to capacity: retaining more PCA dimensions and using a larger kernel training set recovers substantial verification information. At the same time, the required kernel evaluations rise from roughly $2.25\times10^5$ to $5.40\times10^5$ per fold (Table~\ref{tab:cost}). This scaling behaviour motivates the gated architecture. A larger quantum budget improves the specialist but makes universal invocation more expensive; a smaller budget is cheaper but loses more of the ArcFace geometry. Selective routing provides a third axis: rather than choosing only between capacity and cost, the system can reserve the expensive specialist for the subset of pairs where an alternative decision surface may be useful.

\begin{figure}[t]
\centering
\includegraphics[width=.76\textwidth]{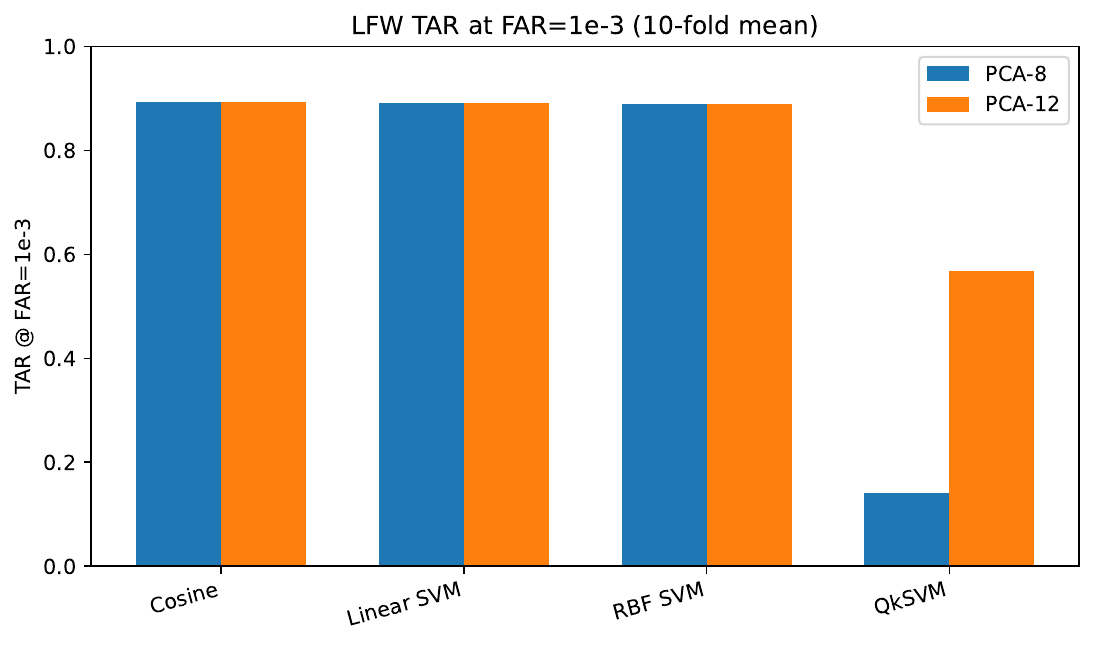}
\caption{Completed LFW qubit-budget study: increasing the compact quantum budget substantially improves TAR at FAR=$10^{-3}$, but the kernel remains constrained by PCA information loss and quadratic evaluation cost.}
\label{fig:lfwbudget}
\end{figure}

\subsection{Cross-Dataset AUC}
\begin{table}[t]
\centering
\caption{Cross-dataset AUC using the strongest reported configuration per method within each completed protocol.}
\label{tab:auc}
\begin{adjustbox}{max width=\textwidth}
\begin{tabular}{lccccc}
\toprule
\textbf{Dataset} & \textbf{Cosine} & \textbf{Full RBF} & \textbf{Matched RBF} & \textbf{QkSVM} & \textbf{Gated Q}\\
\midrule
CASIA-WebFace & \textbf{0.992} & \textbf{0.992} & 0.986 & 0.989 & \textbf{0.992}\\
LFW & \textbf{0.993} & 0.992 & 0.984 & 0.986 & \textbf{0.993}\\
CFP-FP & 0.816 & 0.774 & 0.674 & 0.677 & \textbf{0.824}\\
VGGFace2 & \textbf{0.992} & 0.985 & 0.959 & 0.961 & 0.990\\
\bottomrule
\end{tabular}
\end{adjustbox}
\end{table}

\begin{figure}[t]
\centering
\includegraphics[width=.98\textwidth]{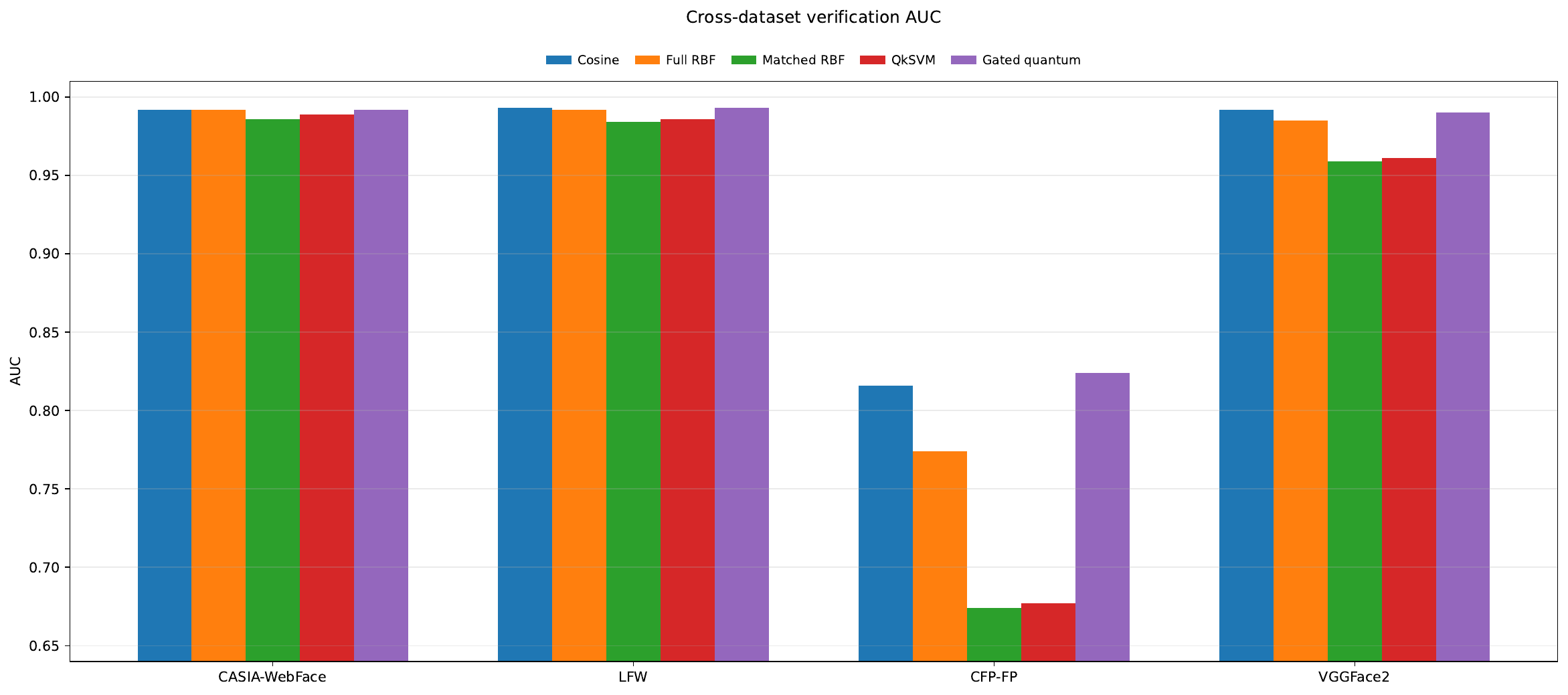}
\caption{Cross-dataset AUC. Gating consistently recovers much of the loss incurred by the budget-constrained QkSVM; the largest benefit is observed on CFP-FP.}
\label{fig:crossauc}
\end{figure}

Table~\ref{tab:auc} and Fig.~\ref{fig:crossauc} demonstrate the main system-level effect of uncertainty gating. The standalone QkSVM underperforms cosine similarity on all evaluated datasets, whereas gating substantially narrows this gap by restricting the specialist to ambiguous pairs and leaving confident cosine decisions unchanged. Relative to the standalone QkSVM, the gated quantum residual improves AUC by approximately 0.003 on CASIA-WebFace, 0.007 on LFW, 0.147 on CFP-FP, and 0.029 on VGGFace2. The largest gain occurs on CFP-FP, indicating that residual specialization is most beneficial when the baseline verifier faces substantial pose-induced ambiguity.

\subsection{CFP-FP: Where Routing Matters Most}
\begin{table}[t]
\centering
\caption{CFP-FP controlled comparison.}
\label{tab:cfp}
\begin{tabular}{lccc}
\toprule
\textbf{Method} & \textbf{AUC} $\uparrow$ & \textbf{EER} $\downarrow$ & \textbf{TAR}@$10^{-2}$ $\uparrow$\\
\midrule
Cosine & 0.816 & 0.258 & 0.206\\
Full RBF & 0.774 & 0.299 & 0.150\\
Matched RBF (KTA) & 0.674 & 0.374 & 0.054\\
QkSVM & 0.677 & 0.373 & 0.054\\
Gated random residual & 0.813 & 0.263 & 0.206\\
Gated classical residual & \textbf{0.824} & 0.252 & 0.223\\
Gated quantum residual & \textbf{0.824} & \textbf{0.251} & \textbf{0.226}\\
\bottomrule
\end{tabular}
\end{table}

\begin{figure}[t]
\centering
\includegraphics[width=.78\textwidth]{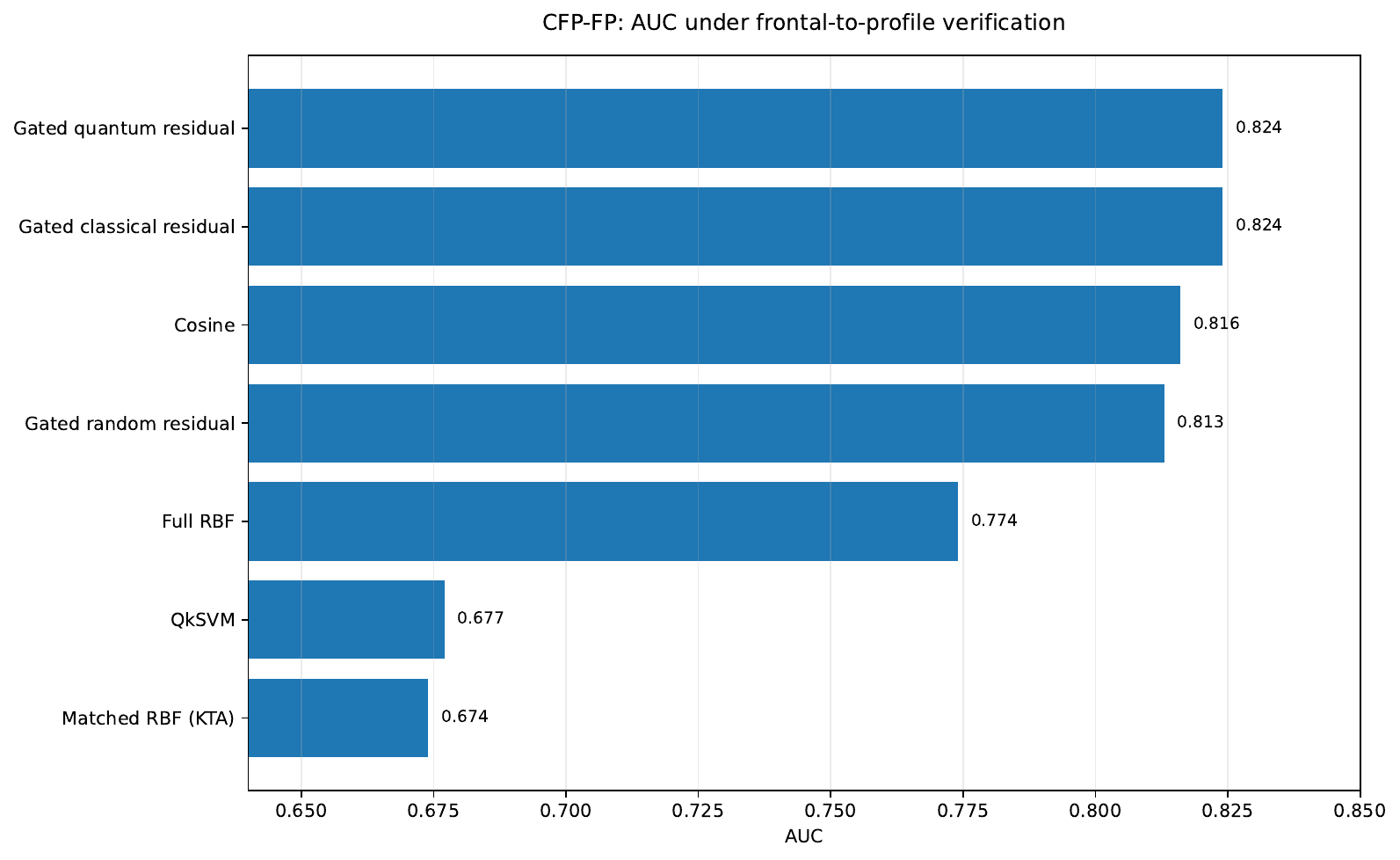}
\caption{CFP-FP AUC. Both gated residuals exceed cosine, while standalone compact kernels remain substantially weaker.}
\label{fig:cfpauc}
\end{figure}

On CFP-FP, the standalone compact models are poor: QkSVM reaches 0.677 AUC, and the matched RBF reaches 0.674 AUC, compared with 0.816 for cosine. Selective routing changes the outcome. The gated quantum residual reaches 0.824 AUC and 0.226 TAR@$10^{-2}$; the identically gated classical residual reaches 0.824 and 0.223. The random residual does not reproduce the gain. This is the most important causal result in the paper. Gating is useful, but the quantum kernel is not uniquely responsible. The correct claim is therefore that \emph{an uncertainty gate converts a globally weak specialist into a useful complementary verifier}; the experiment does not establish quantum advantage.

\begin{figure}[t]
\centering
\includegraphics[width=.78\textwidth]{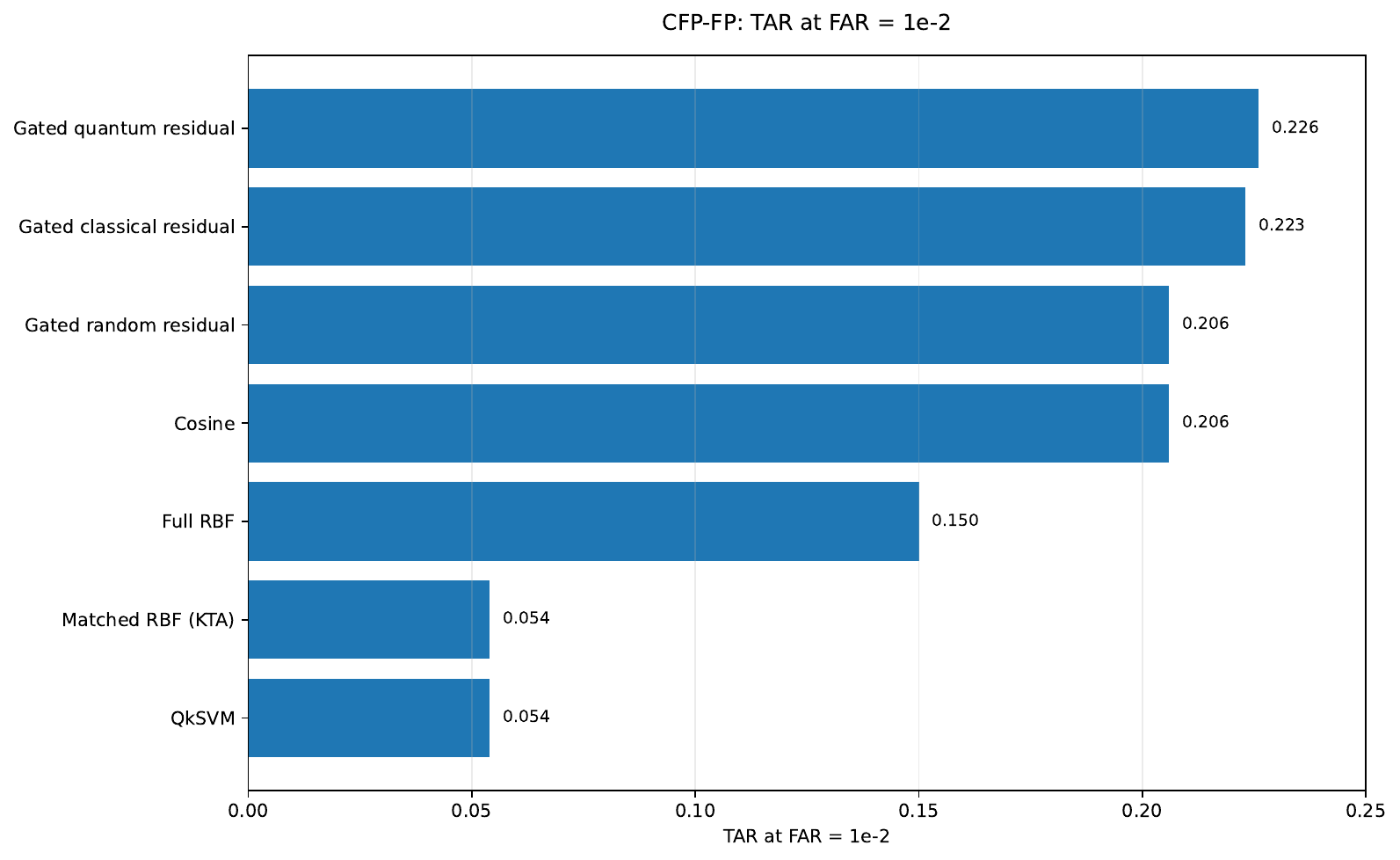}
\caption{CFP-FP TAR at FAR=$10^{-2}$. The quantum residual obtains the highest reported value, but the matched classical residual is extremely close.}
\label{fig:cfptar}
\end{figure}

\subsection{VGGFace2: Protecting a Near-Ceiling Baseline}
On VGGFace2, cosine is already at 0.992 AUC and 0.972 TAR@$10^{-2}$. Standalone QkSVM drops to 0.961/0.589 and matched RBF to 0.959/0.596. The gated quantum residual recovers 0.990/0.969 and the gated classical residual 0.990/0.968. Gating therefore behaves as a safety mechanism: a weak specialist is prevented from degrading the majority of easy pairs.

\begin{table}[t]
\centering
\caption{VGGFace2 controlled comparison.}
\label{tab:vgg}
\begin{tabular}{lccc}
\toprule
Method & AUC $\uparrow$ & EER $\downarrow$ & TAR@$10^{-2}$ $\uparrow$\\
\midrule
Cosine & \textbf{0.992} & \textbf{0.024} & \textbf{0.972}\\
Full RBF & 0.985 & 0.040 & 0.941\\
Matched RBF (KTA) & 0.959 & 0.090 & 0.596\\
QkSVM & 0.961 & 0.084 & 0.589\\
Gated classical residual & 0.990 & 0.027 & 0.968\\
Gated quantum residual & 0.990 & 0.028 & 0.969\\
\bottomrule
\end{tabular}
\end{table}

\begin{figure}[t]
\centering
\includegraphics[width=.86\textwidth]{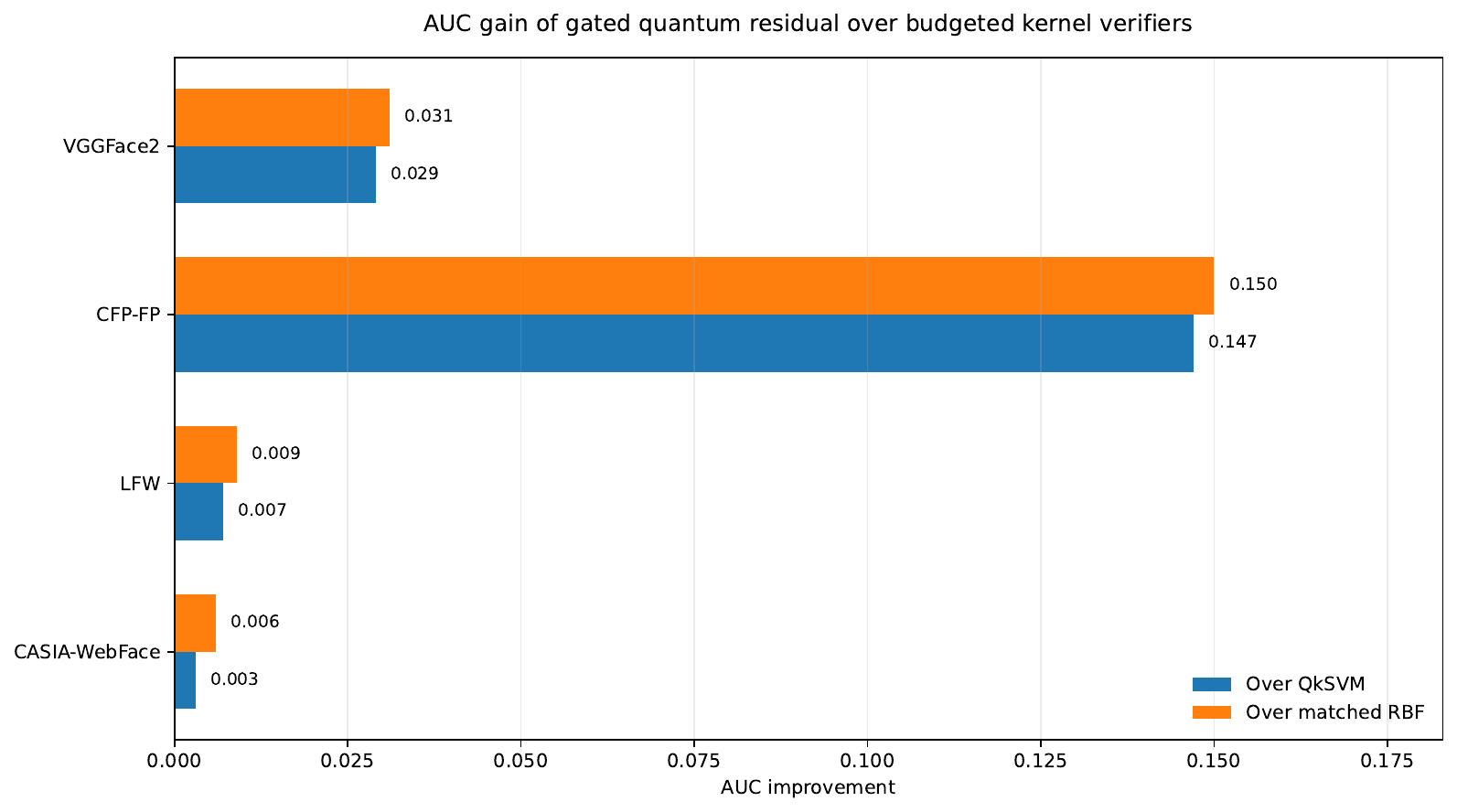}
\caption{AUC gain of the gated quantum residual over standalone QkSVM and the qubit-budgeted RBF control.}
\label{fig:gain}
\end{figure}

\subsection{Cross-Dataset Behavior of Residual Gating}
The gain plot in Fig.~\ref{fig:gain} reveals two distinct regimes. On CASIA-WebFace, LFW, and VGGFace2, uncertainty gating primarily protects the strong cosine baseline from a weaker specialist. On CFP-FP, where severe pose variation creates a larger ambiguous region, residual routing yields a clear improvement over cosine. This suggests that selective specialization is most valuable when the baseline verifier has substantial uncertainty.

\subsection{Control Hierarchy and Attribution}
Table~\ref{tab:controls} states what each control rules out. This hierarchy is central because a hybrid system can improve for reasons unrelated to the quantum feature map. The full RBF tests whether a conventional nonlinear classifier already solves the pair problem when it sees the high-dimensional representation. The matched RBF controls the PCA/qubit input budget. The random residual tests whether changing uncertain scores is sufficient. Finally, the identically gated classical residual isolates the quantum specialist itself while preserving the same routing decision.

\begin{table}[t]
\centering
\scriptsize
\caption{Matched controls and the alternative explanation each address.}
\label{tab:controls}
\begin{tabularx}{\textwidth}{lX}
\toprule
\textbf{Control} & \textbf{What it tests}\\
\midrule
Cosine & Strength of the pretrained face geometry without a learned verifier.\\
Full RBF & Whether generic nonlinear classification helps when compression is removed.\\
Matched RBF & Whether QkSVM gains arise simply from the same low-dimensional bottleneck.\\
Random residual & Whether arbitrary perturbation of uncertain pairs explains gating gains.\\
Gated classical residual & Whether the routing gain is specific to the quantum kernel.\\
\bottomrule
\end{tabularx}
\end{table}

The final comparison is decisive for CFP-FP: gated quantum and gated classical both reach an AUC of 0.824. Thus, the system-level improvement is real, but the available evidence assigns it primarily to selective routing rather than to a uniquely quantum decision geometry. This conservative attribution is important for biometric deployment, where a small apparent gain can otherwise be overstated.

\section{Discussion}
\subsection{Selective Specialization Under Resource Constraints}
The standalone QkSVM is constrained in two ways: PCA compresses a high-quality 512-D face representation to 8 or 12 coordinates, and the kernel itself is estimated over a restricted training budget. A universal replacement strategy, therefore, asks the quantum model to recover information it never receives. Gating avoids this failure mode. High-confidence decisions remain in the representation space for which the backbone was optimized, while only a subset of ambiguous pairs is exposed to the alternative nonlinear similarity. 
This viewpoint also offers a resource argument. If only a fraction of pairs are routed, the expensive quantum kernel need not be invoked universally at deployment. The present simulator study does not measure hardware queue time or energy, so we do not claim a realized resource advantage, but selective invocation is the appropriate systems architecture for testing one.

\subsection{Why the Matched Classical Gate Is Essential}
Without the matched classical residual, the CFP-FP improvement could easily be described as a quantum-kernel gain. That interpretation would be incorrect. The classical residual reaches the same AUC and nearly the same TAR. The shared gate, therefore, explains most of the value. The residual model may still matter for individual samples, but the current evidence does not separate a quantum-specific benefit from that of an equally budgeted classical specialist. This negative attribution result strengthens rather than weakens the paper. It identifies the mechanism that actually survives controlled comparison and prevents a system-level improvement from being misreported as quantum advantage.
\subsection{Biometric Operating Points}
Verification performance must be interpreted at fixed false-acceptance rates, not only by balanced accuracy or AUC. The CFP-FP result is encouraging at FAR=$10^{-2}$, but the experiments do not establish hardware robustness or stringent $10^{-3}$ behaviour for the gated cross-dataset study. Future work should therefore combine finite-shot simulation, calibration, and larger impostor protocols before making deployment claims.

\section{Limitations and Reproducibility}
All quantum kernels are evaluated by analytic statevector simulation. Finite-shot sampling, gate noise, device connectivity, calibration drift, queue latency, and hardware energy are outside the scope of the study. The QkSVM also uses balanced subsets because kernel evaluation is quadratic, whereas classical full baselines can use larger training sets; this reflects near-term resource constraints rather than an accuracy-neutral comparison. The cross-dataset results come from two completed experimental stages rather than one globally shared hyperparameter sweep. We therefore report within-dataset comparisons and avoid interpreting raw AUC differences as dataset difficulty. The matched RBF, gated classical residual, and random residual are included specifically to make causal attribution conservative.

\section{Conclusion}
We evaluated quantum kernels not as replacements for modern face verification but as selectively routed residual specialists. Standalone QkSVM is consistently weaker than cosine because low-qubit compression and kernel-budget constraints discard useful identity geometry. Uncertainty-aware routing substantially reduces this failure: on CFP-FP, the gated quantum residual raises AUC from 0.677 for standalone QkSVM and 0.816 for cosine to 0.824, while improving TAR@$10^{-2}$ to 0.226. On VGGFace2 it preserves near-ceiling cosine performance despite a much weaker standalone specialist. 
However, the identically gated classical residual reaches the same CFP-FP AUC. The supported contribution is therefore a hybrid decision architecture and a methodological lesson: when a compact quantum verifier is globally weaker than a mature biometric score, selective routing can make it useful, but matched classical gating is necessary before any quantum-specific claim is made. This framing offers a realistic path to integrating resource-constrained quantum components into high-performing biometric systems without, by default, sacrificing the strong classical backbone.

\bibliographystyle{splncs04}
\bibliography{mybibliography}
\end{document}